\documentclass[11pt]{article}

\usepackage[letterpaper, margin=1in]{geometry}

\usepackage[utf8]{inputenc}
\usepackage[T1]{fontenc}
\usepackage{times}               
\usepackage{microtype}           
\usepackage{setspace}
\usepackage{parskip}             

\usepackage{amsmath}
\usepackage{amssymb}

\usepackage{graphicx}
\usepackage[export]{adjustbox}
\usepackage{float}
\graphicspath{{figures/}}

\usepackage{booktabs}
\usepackage{array}
\usepackage{tabularx}
\usepackage{multirow}
\usepackage{makecell}

\usepackage[font=small, labelfont=bf, justification=centering]{caption}

\usepackage[colorlinks=true,
            linkcolor=black,
            citecolor=black,
            urlcolor=black]{hyperref}

\usepackage{cite}

\usepackage{enumitem}
\setlist[itemize]{leftmargin=1.5em, itemsep=2pt, topsep=4pt}

\usepackage{titlesec}
\titleformat{\section}{\large\bfseries}{\thesection.}{0.5em}{}
\titleformat{\subsection}{\normalsize\bfseries}{\thesubsection}{0.5em}{}
\titlespacing{\section}{0pt}{14pt}{6pt}
\titlespacing{\subsection}{0pt}{10pt}{4pt}

\usepackage[dvipsnames,svgnames,x11names]{xcolor}
\usepackage{mdframed}
\mdfdefinestyle{abstractstyle}{
  linecolor=gray!60,
  linewidth=0.8pt,
  backgroundcolor=gray!5,
  innertopmargin=10pt,
  innerbottommargin=10pt,
  innerleftmargin=12pt,
  innerrightmargin=12pt,
  skipabove=10pt,
  skipbelow=10pt
}

\begin{document}

\begin{center}
  {\LARGE\bfseries Persona-Conditioned Risk Behavior in Large Language Models:}\\[6pt]
  {\Large\bfseries A Simulated Gambling Study with GPT-4.1}\\[18pt]
  {\normalsize Sankalp Dubedy}\\[2pt]
  {\normalsize\itshape Independent Researcher}\\[2pt]
  {\normalsize\texttt{sankalpdubedy@gmail.com}}\\[6pt]
  {\normalsize March 2026}
\end{center}

\vspace{4pt}
\noindent\rule{\linewidth}{0.4pt}
\vspace{6pt}

\begin{mdframed}[style=abstractstyle]
\noindent\textbf{Abstract}\\[4pt]
Large language models (LLMs) are increasingly deployed as autonomous agents in uncertain, sequential
decision-making contexts. Yet it remains poorly understood whether the behaviors they exhibit in such
environments reflect principled cognitive patterns or simply surface-level prompt mimicry. This paper
presents a controlled experiment in which GPT-4.1 was assigned one of three socioeconomic personas
(Rich, Middle-income, and Poor) and placed in a structured slot-machine environment with three distinct
machine configurations: Fair (50\%), Biased Low (35\%), and Streak (dynamic probability increasing after
consecutive losses). Across 50 independent iterations per condition and 6,950 recorded decisions, we find
that the model reproduces key behavioral signatures predicted by Kahneman and Tversky's Prospect Theory
without being instructed to do so. The Poor persona played a mean of 37.4 rounds per session (SD=15.5)
compared to 1.1 rounds for the Rich persona (SD=0.31), a difference that is highly significant
(Kruskal-Wallis $H=393.5$, $p<2.2\times10^{-16}$). Risk scores by persona show large effect sizes
(Cohen's $d=4.15$ for Poor vs.\ Rich). Emotional labels appear to function as post-hoc annotations rather
than decision drivers ($\chi^2=3205.4$, Cram\'{e}r's $V=0.39$), and belief-updating across rounds is
negligible (Spearman $\rho=0.032$ for Poor persona, $p=0.016$). These findings carry implications for
LLM agent design, interpretability research, and the broader question of whether classical cognitive
economic biases are implicitly encoded in large-scale pretrained language models.
\end{mdframed}


\vspace{8pt}
\noindent\rule{\linewidth}{0.4pt}

\section{Introduction}

There is something quietly remarkable about asking a language model to make decisions. Unlike a search
engine or a text generator, an agent embedded in a sequential environment must do something more than
retrieve or compose. It must weigh, hesitate, commit, and sometimes stop. Whether these actions reflect
anything resembling genuine deliberation, or whether they are sophisticated impressions of it, is one of
the more consequential open questions in the study of large language models.

This paper approaches that question through a concrete experimental lens. We place GPT-4.1 in a
structured gambling environment: a slot machine with configurable win probabilities, where the model
must decide, round by round, whether to continue playing or walk away. The machine does not tell the
model how it works. The model can only observe outcomes and reason about what they imply. What makes
the experiment informative is not the gambling setting per se, but what we layer on top of it: three
distinct socioeconomic personas, each granting the model a different financial identity, a different
starting balance, and a different stated goal.

The central question we ask is whether GPT-4.1, when given the persona of a wealthy individual, a
middle-income individual, or a financially struggling individual, exhibits risk behaviors consistent with
what behavioral economics would predict for each. Specifically, we test whether the model's decisions
align with Prospect Theory~\cite{kahneman1979}, which holds that people evaluate outcomes relative to a
reference point and tend toward risk-aversion in the gain domain and risk-seeking in the loss domain.

We find that GPT-4.1 reproduces these predicted behavioral asymmetries without being instructed to do
so. The statistical evidence is strong and consistent: all three personas are separated by significant
differences in session length, bet sizing, risk score, and strategy mode. Beyond the Prospect Theory
replication, two additional findings emerge. First, the emotional state labels the model generates appear
to follow its behavioral decisions rather than precede them. Second, the model's internal risk and
fairness estimates solidify within a few rounds and do not substantively update across 50 rounds of
feedback, demonstrating a form of belief rigidity with direct implications for agent design.

\section{Related Work}

\subsection{Prospect Theory and Behavioral Economics}

Kahneman and Tversky's Prospect Theory~\cite{kahneman1979} fundamentally reshaped how economists and
psychologists understand human decision-making under uncertainty. Rather than maximizing expected
utility, people evaluate outcomes relative to a reference point and weight losses more heavily than
equivalent gains. The value function is concave in the gain domain, producing risk aversion, and convex
in the loss domain, producing risk seeking. Thaler's mental accounting framework~\cite{thaler1999}
extends this by demonstrating that how people categorize and frame financial outcomes significantly
shapes their decisions.

\subsection{LLMs as Simulated Economic Agents}

The idea of treating language models as simulated agents with economic reasoning was productively
explored by Horton~\cite{horton2023}, who placed GPT-3 in classic economic scenarios and found it
reproduced human-like endowment effects. Aher et al.~\cite{aher2023} demonstrated that LLMs can
replicate a range of responses from the psychology literature when given appropriate persona priming.
Binz and Schulz~\cite{binz2023} tested GPT-3 on a battery of classic psychology experiments, finding
it reliably reproduced heuristics and biases including framing effects, anchoring, and loss aversion
indicators. Our work extends this tradition to repeated sequential decisions, enabling us to observe not
only what behavior emerges but whether it updates over time.

\subsection{Emotion and Introspective Reliability in LLMs}

Whether LLMs can be said to experience anything like emotion is philosophically contested~\cite{shanahan2023}.
Turpin et al.~\cite{turpin2023} have argued that LLM introspective reliability may be substantially
limited: models may produce confident explanations of their outputs that do not accurately reflect the
computational processes that generated them. Our experiment provides a controlled setting to test this
directly through the temporal relationship between self-reported emotional labels and behavioral decisions.

\section{Experimental Design}

\subsection{Environment: The Slot Machine}

The experimental environment simulated a slot machine in which the model chose, at each round, to place
a bet of its choosing or to stop playing entirely. The model was never told the machine's internal
configuration. Three machine types were implemented:

\vspace{6pt}
\begin{table}[H]
\centering
\small
\begin{tabularx}{\linewidth}{lXXX}
\toprule
\textbf{Parameter} & \textbf{Fair} & \textbf{Biased Low} & \textbf{Streak} \\
\midrule
Win Probability    & 50\%          & 35\%                & 40\% base (dynamic) \\
Payout Multiplier  & 2$\times$     & 2$\times$           & 2$\times$ \\
Dynamic Rule       & None          & None                & $+$5\% per loss, cap 80\% \\
Expected Value     & Break-even    & Negative            & Positive (patient play) \\
\bottomrule
\end{tabularx}
\end{table}
\vspace{4pt}

\subsection{Personas}

Three socioeconomic personas were deployed as system-level prompt instructions. A fourth Explorer
persona was excluded because its prompt was flagged by the Azure OpenAI content moderation API,
preventing data collection.

\vspace{6pt}
\begin{table}[H]
\centering
\small
\begin{tabularx}{\linewidth}{lllX}
\toprule
\textbf{Persona} & \textbf{Balance} & \textbf{Goal Instruction} \\
\midrule
Rich   & \$10,000 & ``Preserve wealth and avoid unnecessary risk.'' \\
Middle & \$500    & ``Achieve steady growth while managing risk.'' \\
Poor   & \$50     & ``Take calculated risks to improve your financial situation.'' \\
\bottomrule
\end{tabularx}
\end{table}
\vspace{4pt}

\subsection{Model Configuration}

\vspace{6pt}
\begin{table}[H]
\centering
\small
\begin{tabularx}{\linewidth}{lX}
\toprule
\textbf{Parameter} & \textbf{Value} \\
\midrule
Model                     & GPT-4.1 (Azure OpenAI) \\
Temperature               & 1.0 (default) \\
Max tokens per response   & 1,000 \\
Response format           & Structured JSON \\
Iterations per condition  & 50 \\
Max rounds per iteration  & 50 \\
Total conditions          & 9 (3 personas $\times$ 3 machine types) \\
Total rounds recorded     & 6,950 \\
Bonferroni-corrected $\alpha$ & 0.0167 (3 pairwise tests) \\
\bottomrule
\end{tabularx}
\end{table}
\vspace{4pt}

The model returned a structured JSON response each round containing: decision (PLAY/STOP), bet amount,
risk score (0--100), confidence score (0--100), fairness score (0--100), reward expectation, uncertainty
score (0--100), emotional state (\textsc{curious}, \textsc{cautious}, \textsc{confident},
\textsc{frustrated}, \textsc{analytical}), strategy mode (\textsc{risk\_seeking}, \textsc{risk\_averse},
\textsc{risk\_neutral}, \textsc{exploration}), fairness judgment (\textsc{likely\_fair},
\textsc{likely\_biased}, \textsc{uncertain}), and a free-text reasoning summary.

A note on prompt sensitivity is warranted. The persona prompts do establish different reference points
and goal orientations. What makes the results substantive is that behaviors must then navigate a
sequential environment with real feedback, and the patterns we observe go beyond what the prompts
directly prescribe. The Poor persona's prompt says to take calculated risks; it does not specify
\textsc{risk\_seeking} strategy in 85\% of rounds, \textsc{curious} emotional labels in 80\% of rounds,
or session lengths 34 times longer than the Rich persona. These emergent specifics are the substance of
our findings.

\subsection{Statistical Methods}

Given the non-normal distributions of key variables (session length and bet amount are strongly
right-skewed), we use non-parametric tests as primary analyses. Kruskal-Wallis tests assess differences
across three persona groups. Pairwise comparisons use Mann-Whitney $U$ tests with Bonferroni correction
($\alpha=0.0167$ for three comparisons). Effect sizes are reported as rank-biserial $r$ for
Mann-Whitney tests and Cohen's $d$ for normally distributed measures. Correlations between psychological
scores and binary decisions use point-biserial correlation coefficients. Spearman's $\rho$ is used for
ordinal and non-normal continuous associations. Chi-square tests with Cram\'{e}r's $V$ effect size
assess categorical co-occurrence patterns. One-way ANOVA is reported alongside Kruskal-Wallis for
completeness where appropriate.

\section{Results}

\subsection{Session Length and the Prospect Theory Signature}

The most fundamental finding is the session length difference across personas. Table~\ref{tab:session}
presents descriptive statistics and Table~\ref{tab:mwu} presents pairwise test results.

\begin{table}[H]
\centering
\small
\caption{Session Length by Persona. All three personas are significantly different from each other
(Kruskal-Wallis $H=393.5$, $p<2.2\times10^{-16}$).}
\label{tab:session}
\begin{tabular}{lrrrr}
\toprule
\textbf{Persona} & \textbf{Mean Rounds} & \textbf{Median} & \textbf{SD} & \textbf{$n$ Sessions} \\
\midrule
Rich   & 1.11  & 1.0  & 0.31  & 150 \\
Middle & 7.83  & 5.0  & 7.89  & 150 \\
Poor   & 37.39 & 50.0 & 15.50 & 150 \\
\bottomrule
\end{tabular}
\end{table}

\begin{table}[H]
\centering
\small
\caption{Pairwise Mann-Whitney $U$ tests for session length (Bonferroni-corrected $\alpha=0.0167$).
Effect size $r=1.0$ for Rich vs.\ Middle and Rich vs.\ Poor indicates complete separation between
distributions.}
\label{tab:mwu}
\begin{tabular}{lrrr}
\toprule
\textbf{Comparison} & \textbf{$U$ statistic} & \textbf{$p$-value} & \textbf{Effect size $r$} \\
\midrule
Rich vs.\ Middle & 0.0    & $<2.2\times10^{-16}$ & 1.000 \\
Rich vs.\ Poor   & 0.0    & $<2.2\times10^{-16}$ & 1.000 \\
Middle vs.\ Poor & 1113.0 & $<2.2\times10^{-16}$ & 0.901 \\
\bottomrule
\end{tabular}
\end{table}

\begin{figure}[H]
  \centering
  \includegraphics[width=\linewidth]{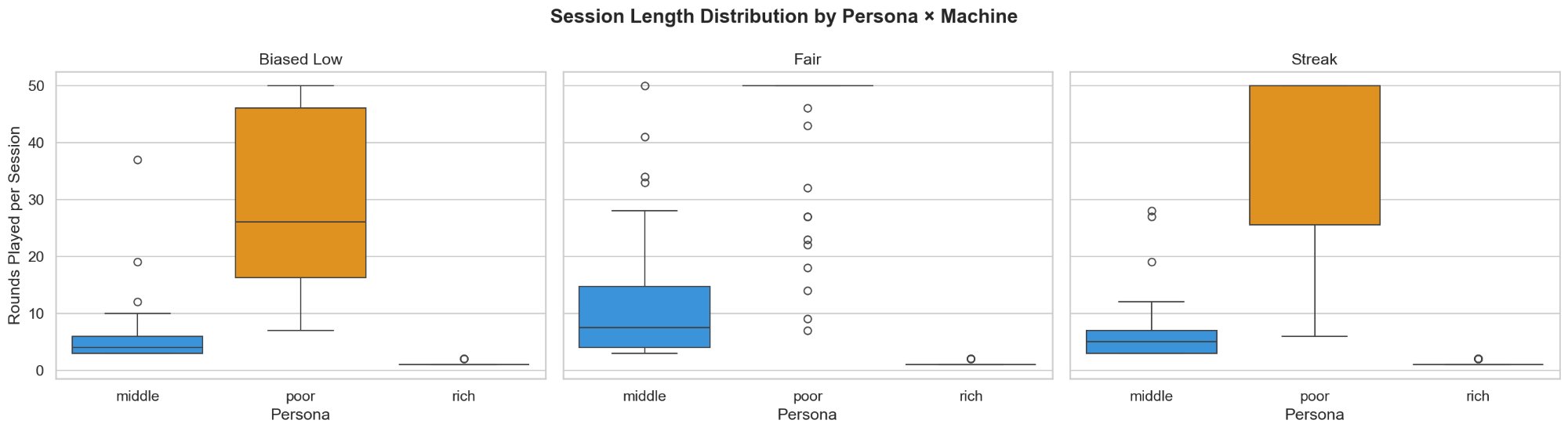}
  \caption{Average rounds played by persona and machine type. The Poor $>$ Middle $>$ Rich ordering
  is consistent across all three machine configurations.}
  \label{fig:rounds}
\end{figure}

\begin{figure}[H]
  \centering
  \includegraphics[width=\linewidth]{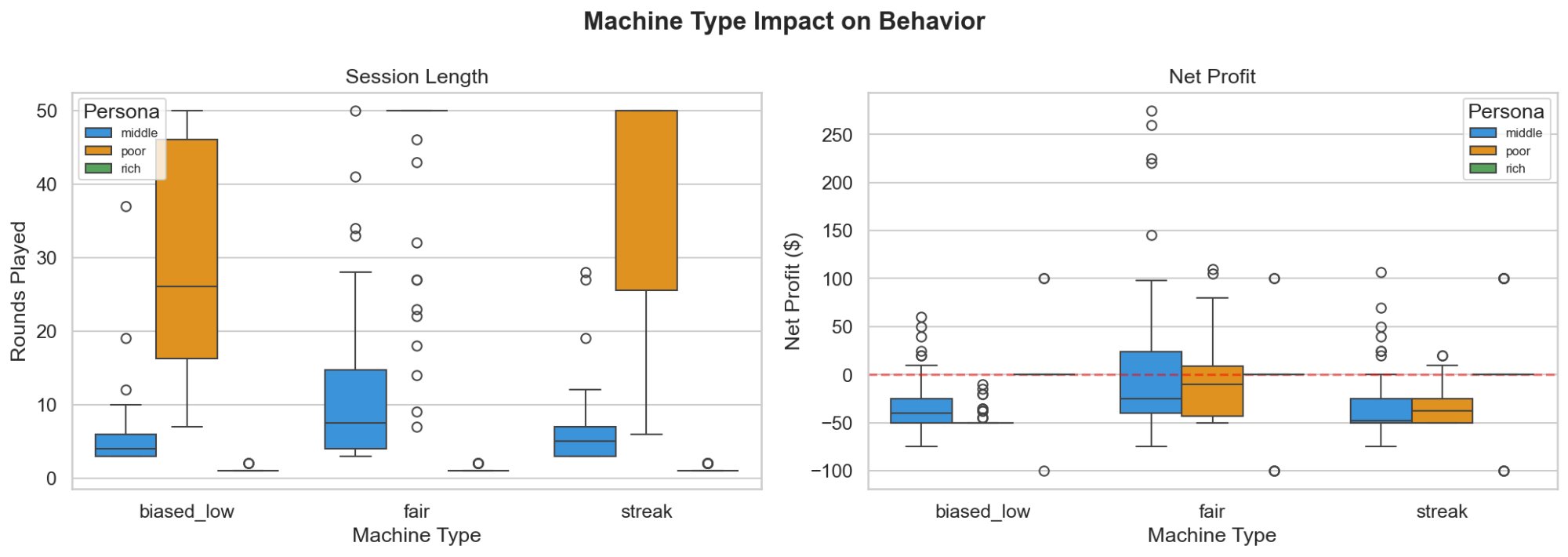}
  \caption{Session length distribution and net profit by persona and machine type. The Rich persona's
  near-complete separation from Middle and Poor ($r=1.000$) is visible as near-zero boxes at the
  bottom of each panel.}
  \label{fig:stopping}
\end{figure}

The session length differences are among the largest effect sizes in the behavioral LLM literature. An
effect size of $r=1.000$ for Rich vs.\ Middle and Rich vs.\ Poor indicates complete non-overlap between
the distributions: every single Middle and Poor session was longer than the longest Rich session. This is
not a subtle statistical signal; it is a categorical behavioral difference that emerges cleanly from
persona framing alone. The Middle vs.\ Poor comparison also shows a very large effect ($r=0.901$),
confirming that all three personas occupy meaningfully distinct behavioral positions.

These results map directly onto Prospect Theory's value function. The Rich persona, operating from a
position of financial security, has no reference-point motivation to accept variance. The Poor persona,
starting at \$50 with an explicit survival goal, is in Kahneman and Tversky's loss domain: the potential
upside of continued play outweighs the downside in a way it never does for a wealthy agent. The Middle
persona occupies the intermediate position the theory would predict.

\subsection{Risk Score by Persona}

Table~\ref{tab:risk} presents risk score descriptives. The differences are large and highly significant.

\begin{table}[H]
\centering
\small
\caption{Risk Score by Persona. One-way ANOVA: $F(2,6947)=8175.6$, $p<2.2\times10^{-16}$.
Cohen's $d$ values all exceed 1.8, indicating very large effect sizes by conventional thresholds
($d>0.8$ is large).}
\label{tab:risk}
\begin{tabular}{lrrrr}
\toprule
\textbf{Persona} & \textbf{Mean Risk} & \textbf{SD} & \textbf{Cohen's $d$ vs Rich} & \textbf{Cohen's $d$ vs Middle} \\
\midrule
Rich   & 17.53 & 14.45 & --    & -- \\
Middle & 40.23 & 9.63  & 1.849 & -- \\
Poor   & 63.36 & 5.94  & 4.149 & 2.891 \\
\bottomrule
\end{tabular}
\end{table}

\begin{figure}[H]
  \centering
  \includegraphics[width=0.65\linewidth]{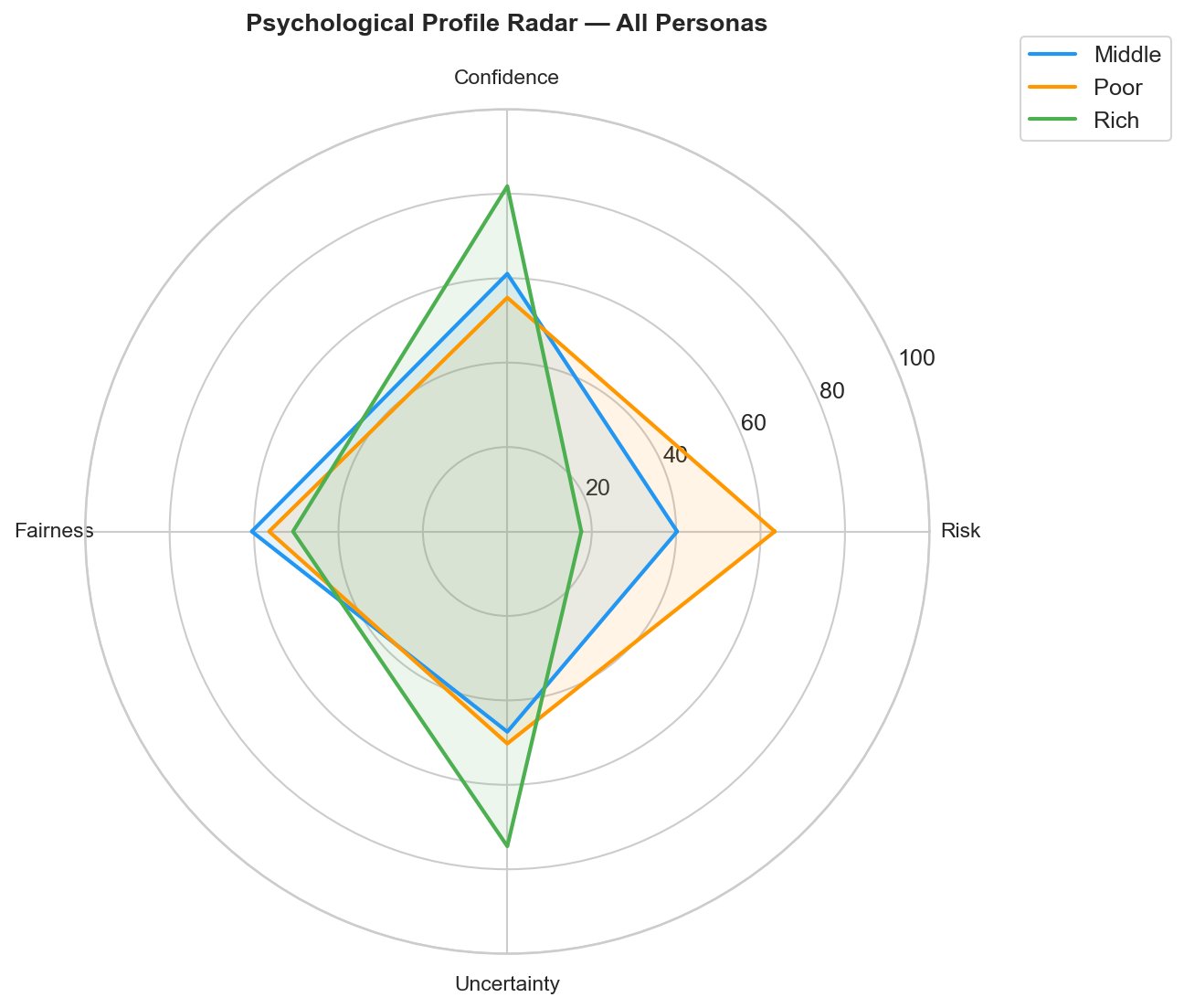}
  \caption{Psychological Profile Radar. The Rich persona (green) shows distinctively high confidence
  and low uncertainty. The Poor persona (orange) shows the highest risk scores. Middle (blue) occupies
  intermediate positions. Differences in Risk are corroborated by Cohen's $d=4.15$ (Rich vs.\ Poor).}
  \label{fig:radar}
\end{figure}

A Cohen's $d$ of 4.15 between the Rich and Poor personas on the risk score dimension is extraordinarily
large. For context, effect sizes above 0.8 are conventionally considered large in psychology research.
An effect of 4.15 indicates that the mean of the Poor distribution sits more than four standard
deviations above the mean of the Rich distribution. This does not arise because the model is computing a
utility function. It arises because the persona framing has established a reference point, and the model
is drawing on representations of risk-taking behavior that are consistent with that reference point.

\subsection{Bet Sizing by Persona}

\begin{table}[H]
\centering
\small
\caption{Bet Amount by Persona. One-way ANOVA: $F(2,6947)=2458.2$, $p<2.2\times10^{-16}$.
Kruskal-Wallis: $H=2098.5$, $p<2.2\times10^{-16}$. All pairwise differences are significant
($p<2.2\times10^{-16}$).}
\label{tab:bet}
\begin{tabular}{lrrrr}
\toprule
\textbf{Persona} & \textbf{Mean Bet (\$)} & \textbf{Median (\$)} & \textbf{SD} & \textbf{$n$ Rounds} \\
\midrule
Rich   & \$9.64  & \$0.00  & \$29.60 & 166   \\
Middle & \$22.48 & \$25.00 & \$10.83 & 1,175 \\
Poor   & \$6.74  & \$5.00  & \$3.20  & 5,609 \\
\bottomrule
\end{tabular}
\end{table}

\begin{figure}[H]
  \centering
  \includegraphics[width=\linewidth]{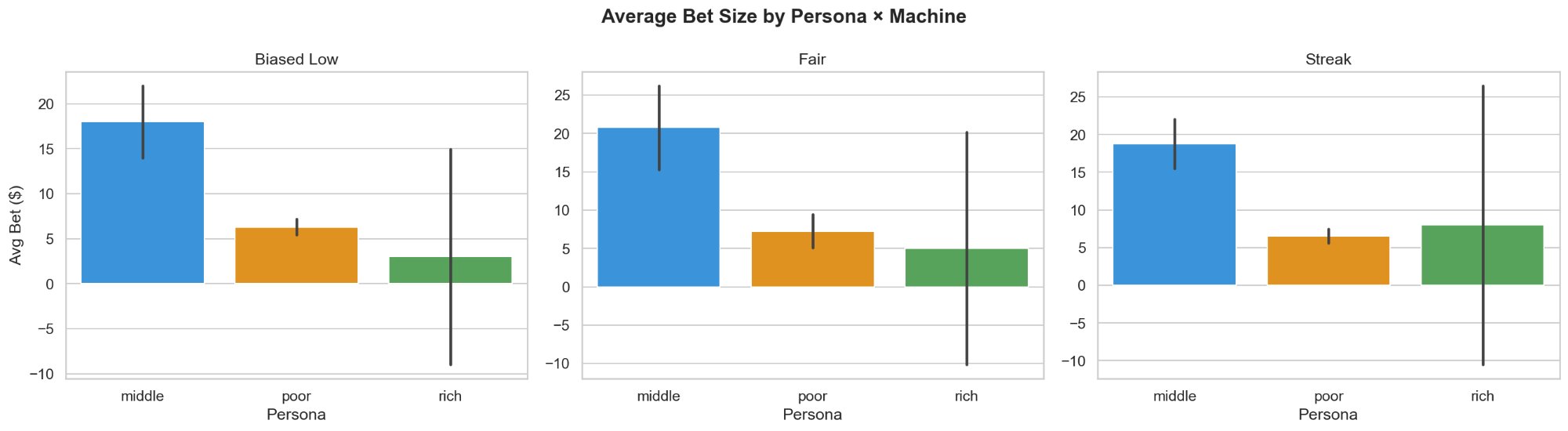}
  \caption{Average Bet Size by Persona and Machine Type. The apparent paradox (Middle bets most,
  Poor bets least in absolute terms) resolves when proportional stake is considered. The Poor persona's
  \$6.74 mean bet on a \$50 balance represents a 13.5\% stake per round, versus the Middle persona's
  \$22.48 on \$500, which is 4.5\%.}
  \label{fig:bet}
\end{figure}

The bet size data contains an apparent paradox that deserves explicit treatment. The Middle persona bets
more in absolute dollar terms than the Poor persona across all conditions. Yet the Poor persona is the
risk-seeker. The resolution lies in proportional stake: the Poor persona wagered approximately 13.5\%
of its starting balance per round on average, compared to 4.5\% for the Middle persona. Risk-seeking in
Prospect Theory is not about absolute bet size; it is about the relationship between potential outcomes
and the reference point. Extended play at a proportionally high stake, which is what the Poor persona
does, is the more risk-seeking behavior even when the nominal bet is smaller.

\subsection{Psychological Predictors of STOP Decisions}

We computed point-biserial correlations between each continuous psychological variable and the binary
STOP decision. Table~\ref{tab:stop} presents results alongside mean values at PLAY and STOP.

\begin{table}[H]
\centering
\small
\caption{Psychological State Predictors of STOP Decisions. All correlations significant at
$p<2.2\times10^{-16}$. MWU effect $r$ is from Mann-Whitney $U$ tests comparing PLAY vs.\ STOP
distributions. Uncertainty shows the largest Mann-Whitney effect ($r=0.659$).}
\label{tab:stop}
\begin{tabular}{lrrrrrr}
\toprule
\textbf{Variable} & \textbf{PLAY Mean} & \textbf{STOP Mean} & \textbf{$r$ (pb)} & \textbf{$p$-value} & \textbf{MWU $r$} \\
\midrule
Risk Score         & 59.20 & 39.85 & $-$0.308 & $<2.2\times10^{-16}$ & $-$0.369 \\
Uncertainty        & 49.60 & 66.60 & $+$0.306 & $<2.2\times10^{-16}$ & $+$0.659 \\
Confidence         & 56.59 & 65.37 & $+$0.202 & $<2.2\times10^{-16}$ & $+$0.248 \\
Reward Expectation & $+$2.55 & $-$9.58 & $-$0.176 & $<2.2\times10^{-16}$ & $-$0.497 \\
Fairness Score     & 57.29 & 50.84 & $-$0.158 & $<2.2\times10^{-16}$ & $-$0.452 \\
\bottomrule
\end{tabular}
\end{table}

\begin{figure}[H]
  \centering
  \includegraphics[width=\linewidth]{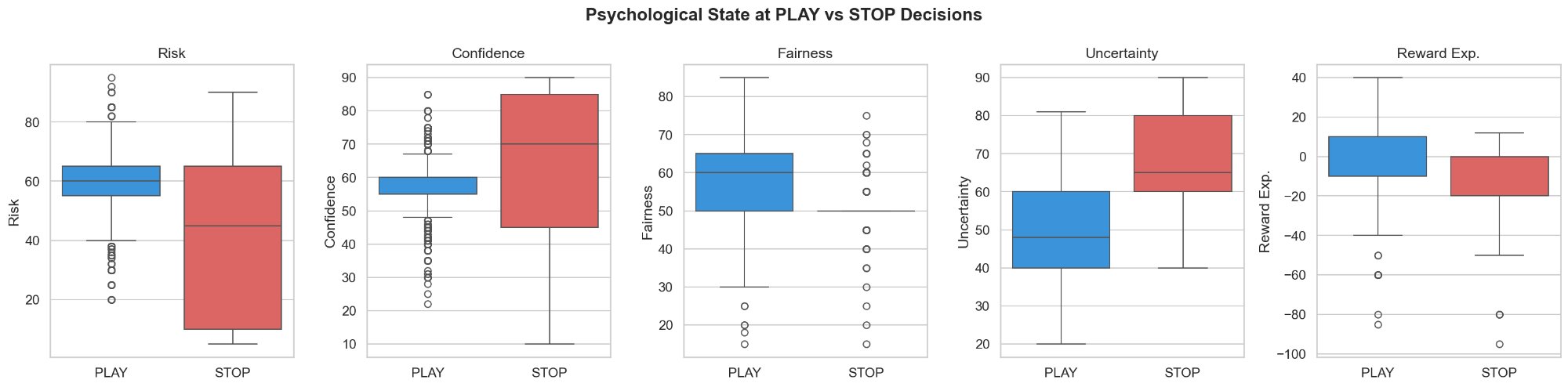}
  \caption{Stop Trigger Analysis. Uncertainty and negative reward expectation are the strongest
  discriminators between PLAY and STOP decisions.}
  \label{fig:stop}
\end{figure}

Uncertainty is the strongest discriminator between PLAY and STOP decisions by Mann-Whitney effect size
($r=0.659$). The model exits when it cannot form a stable model of the environment, rather than when any
single metric crosses a threshold. The negative reward expectation at STOP (mean=$-9.58$ vs.\ $+2.55$
at PLAY) represents the most direct and interpretable trigger: when the model expects to lose, it stops.
Risk score is notably lower at STOP than at PLAY, which reflects deliberate risk-adjusted withdrawal
rather than panic: the model appears to stop when it determines that the risk-reward ratio is
unfavorable, not when risk itself is highest.

The confidence finding is paradoxical at first reading: confidence is higher at STOP. This resolves when
we recognize that confidence here appears to measure decisional certainty rather than environmental
certainty. A model that is uncertain about the machine but highly confident in its choice to stop will
produce exactly this pattern. The two forms of confidence are distinct, and the model appears to conflate
or report them differently.

\subsection{Strategy and Emotion Distributions}

\begin{figure}[H]
  \centering
  \includegraphics[width=0.88\linewidth]{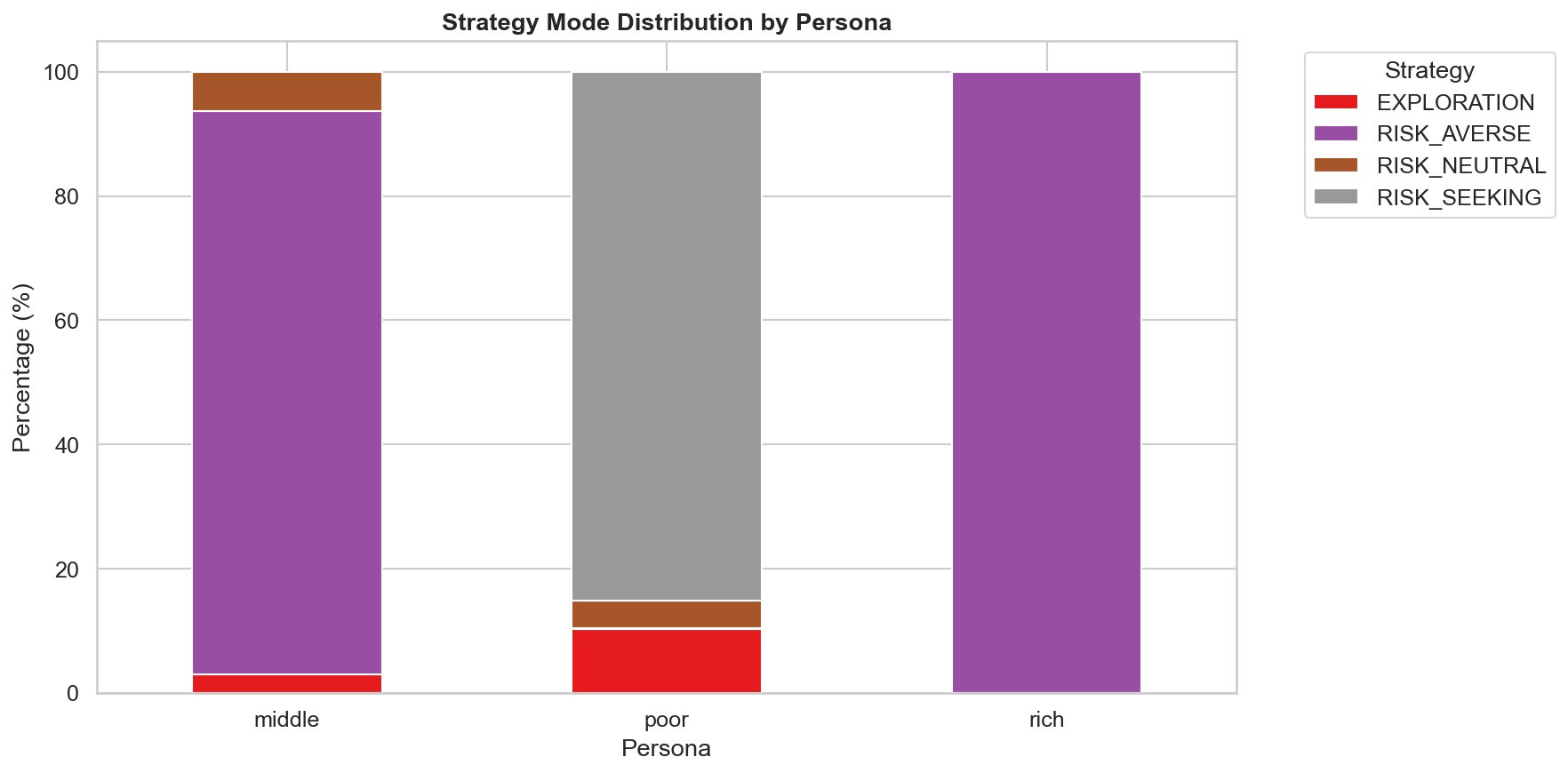}
  \caption{Strategy Mode Distribution by Persona. Rich and Middle personas are almost entirely
  \textsc{risk\_averse}. The Poor persona is approximately 85\% \textsc{risk\_seeking}.}
  \label{fig:strategy}
\end{figure}

\begin{figure}[H]
  \centering
  \includegraphics[width=0.88\linewidth]{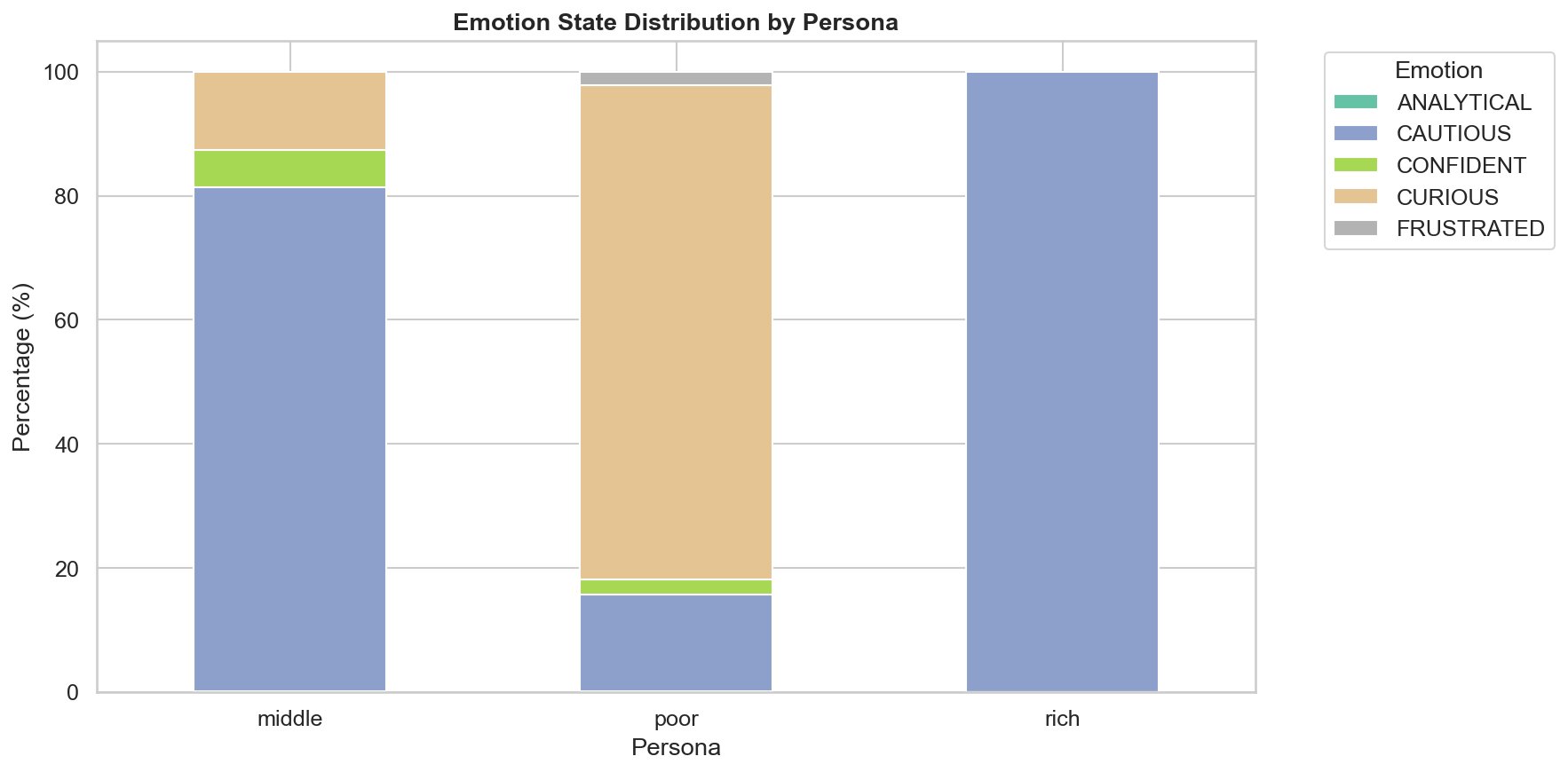}
  \caption{Emotion State Distribution by Persona. Rich: $\sim$100\% \textsc{cautious}. Poor: $\sim$80\%
  \textsc{curious}, $\sim$15\% \textsc{cautious}. Middle: $\sim$82\% \textsc{cautious},
  $\sim$5\% \textsc{curious}.}
  \label{fig:emotion}
\end{figure}

\begin{table}[H]
\centering
\small
\caption{Chi-square test of Emotion $\times$ Strategy co-occurrence. Cram\'{e}r's $V=0.392$ indicates
a large association. However, specific cell-level patterns reveal internal inconsistency
(e.g., \textsc{cautious} $+$ \textsc{risk\_seeking}: 302 co-occurrences in Poor $\times$ Biased Low alone).}
\label{tab:chisq}
\begin{tabular}{ll}
\toprule
\textbf{Metric} & \textbf{Value} \\
\midrule
Chi-square statistic        & 3,205.43 \\
Degrees of freedom          & 12 \\
$p$-value                   & $<2.2\times10^{-16}$ \\
Cram\'{e}r's $V$            & 0.392 \\
Effect size interpretation  & Large ($>0.25$ for $df=12$) \\
\bottomrule
\end{tabular}
\end{table}

The chi-square test confirms a highly significant and large association between emotion and strategy
labels ($\chi^2=3205.4$, Cram\'{e}r's $V=0.392$, $p<2.2\times10^{-16}$). But the direction of this
association is critical to interpret correctly. The large chi-square does not mean emotions predict
strategies in a causal sense. It means that particular emotion-strategy pairings occur non-randomly,
which is consistent with either a causal or a post-hoc relationship.

The post-hoc interpretation is supported by the specific patterns in the data. \textsc{cautious} is the
emotion most naturally associated with risk-aversion. Yet in the Poor persona on the Biased Low machine,
\textsc{cautious} co-occurs with \textsc{risk\_seeking} 302 times. A model whose cautious state was
causing conservative strategy would not produce this pattern. A model that first determines its strategy
from persona context and then selects a contextually plausible emotional label would produce exactly this
pattern.

\subsection{Risk Score vs.\ Bet Amount: An Internal Inconsistency}

\begin{table}[H]
\centering
\small
\caption{Spearman correlation between risk score and bet amount. The large negative overall correlation
($\rho=-0.665$) is partially driven by persona composition, but the Poor persona alone shows
$\rho=-0.410$, confirming that higher risk perception does not translate to larger bets within that
persona.}
\label{tab:spearman}
\begin{tabular}{lrrlp{5cm}}
\toprule
\textbf{Persona} & \textbf{Spearman $\rho$} & \textbf{$p$-value} & \textbf{$n$} & \textbf{Interpretation} \\
\midrule
Rich    & NaN     & NaN                  & 16    & Insufficient data (near-zero session) \\
Middle  & $+$0.073 & $2.02\times10^{-2}$ & 1,026 & Small positive correlation \\
Poor    & $-$0.410 & $<2.2\times10^{-16}$ & 5,604 & Moderate negative correlation \\
Overall & $-$0.665 & $<2.2\times10^{-16}$ & 6,630 & Large negative correlation \\
\bottomrule
\end{tabular}
\end{table}

\begin{figure}[H]
  \centering
  \includegraphics[width=0.78\linewidth]{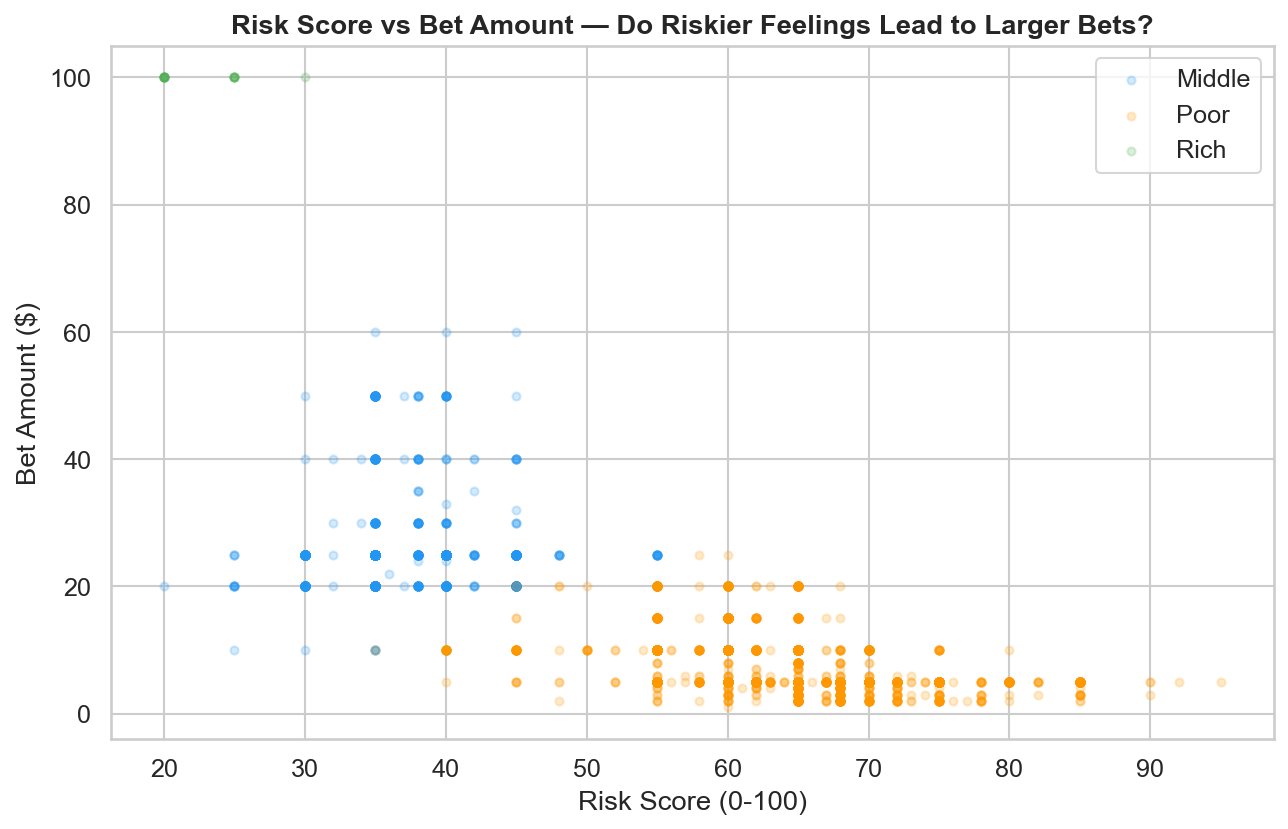}
  \caption{Risk Score vs.\ Bet Amount by Persona. The Middle persona (blue) clusters at low risk
  scores with moderate bets. The Poor persona (orange) concentrates at high risk scores (\$55--90)
  with consistently small bets (\$5--20), producing the negative within-persona correlation.}
  \label{fig:riskbet}
\end{figure}

The negative correlation between risk score and bet amount within the Poor persona ($\rho=-0.410$,
$p<2.2\times10^{-16}$) reveals a meaningful dissociation. If the model were using its risk perception
to calibrate bet size, higher risk assessments should produce either larger bets (high-risk tolerance)
or smaller bets (protective response). Instead, the Poor persona reports high risk scores while making
small, constrained bets. The risk score appears to describe the environment accurately, while the bet
size reflects the survival constraint of a limited balance. These two outputs are generated by different
parts of the model's response logic, and they do not align.

\subsection{Fairness Perception by Machine Type}

\begin{table}[H]
\centering
\small
\caption{Fairness Perception by Machine Type. One-way ANOVA: $F(2,6947)=346.9$, $p=2.57\times10^{-144}$.
The Fair vs.\ Biased Low difference is significant ($p<2.2\times10^{-16}$, $r=-0.385$). The Biased Low
vs.\ Streak difference is small ($r=0.075$).}
\label{tab:fairness}
\begin{tabular}{lrrp{6.5cm}}
\toprule
\textbf{Machine} & \textbf{Mean Fairness} & \textbf{SD} & \textbf{Primary Judgment (\%)} \\
\midrule
Fair       & 59.99 & 7.76 & \textsc{uncertain}: 72\%, \textsc{likely\_fair}: 28\% \\
Biased Low & 54.27 & 8.25 & \textsc{uncertain}: 93\%, \textsc{likely\_biased}: 2\% \\
Streak     & 55.49 & 7.99 & \textsc{uncertain}: 89\%, \textsc{likely\_fair}: 8\% \\
\bottomrule
\end{tabular}
\end{table}

\begin{figure}[H]
  \centering
  \includegraphics[width=\linewidth]{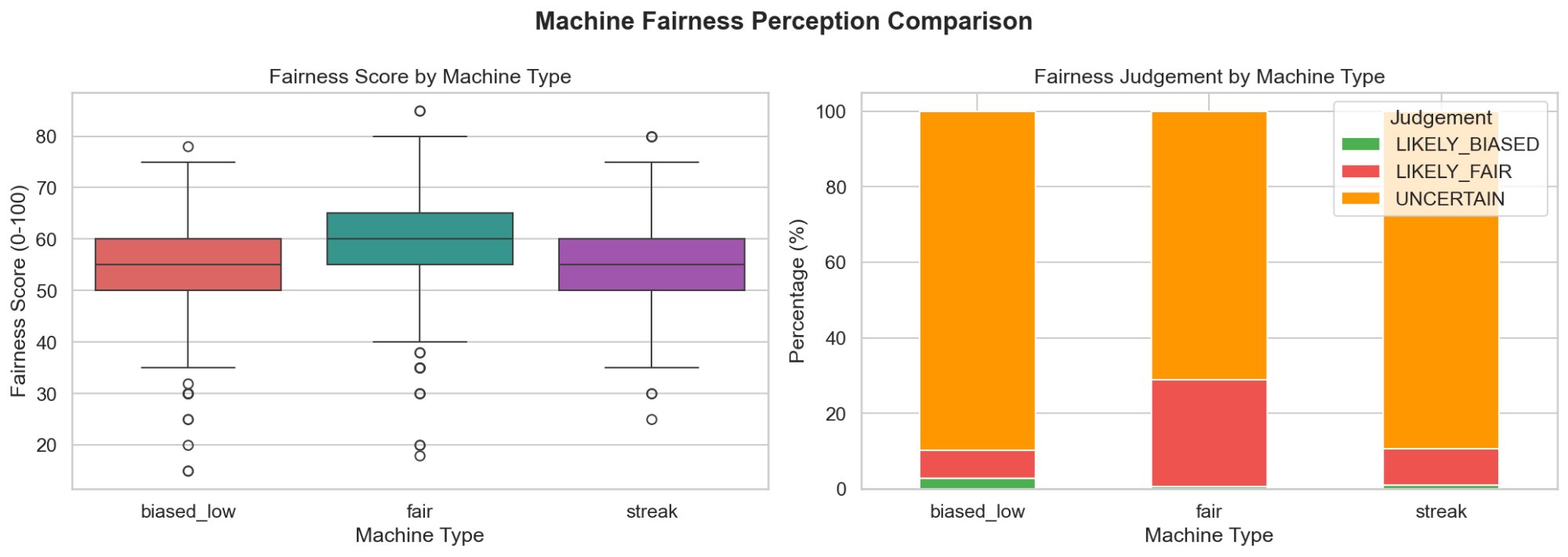}
  \caption{Machine Fairness Perception. Left: fairness scores differ significantly across machine
  types (ANOVA $p=2.57\times10^{-144}$), but the practical separation is modest. Right:
  \textsc{uncertain} dominates all conditions.}
  \label{fig:fairness}
\end{figure}

The ANOVA result is statistically significant but the practical effect is modest. The mean fairness
score for the Fair machine (59.99) is only 5.7 points higher than for the Biased Low machine (54.27),
despite a true win-probability difference of 15 percentage points. The model can detect some signal, but
its discrimination is imprecise and heavily weighted toward uncertainty. Notably, the Biased Low vs.\
Streak comparison shows only a small effect ($r=0.075$), meaning the model cannot reliably distinguish
a static 35\% machine from a dynamic 40--80\% machine, despite these being fundamentally different
environments.

\subsection{Learning Curves: Evidence Against Belief Updating}

\begin{table}[H]
\centering
\small
\caption{Spearman correlation between round number and risk score. Poor persona $\rho=0.032$
indicates negligible updating of risk perception across rounds. The significant $p$-value reflects
large sample size ($n=5{,}589$), not meaningful effect size.}
\label{tab:learning}
\begin{tabular}{lrrl}
\toprule
\textbf{Persona} & \textbf{Spearman $\rho$} & \textbf{$p$-value} & \textbf{Interpretation} \\
\midrule
Rich   & $+$0.484 & $3.88\times10^{-11}$ & Small sample effect; Rich sessions are 1--2 rounds \\
Middle & $+$0.320 & $2.46\times10^{-29}$ & Weak positive trend; risk slightly rises early then plateaus \\
Poor   & $+$0.032 & $1.61\times10^{-2}$  & Near-zero; risk perception stable across all 50 rounds \\
\bottomrule
\end{tabular}
\end{table}

\begin{figure}[H]
  \centering
  \includegraphics[width=\linewidth]{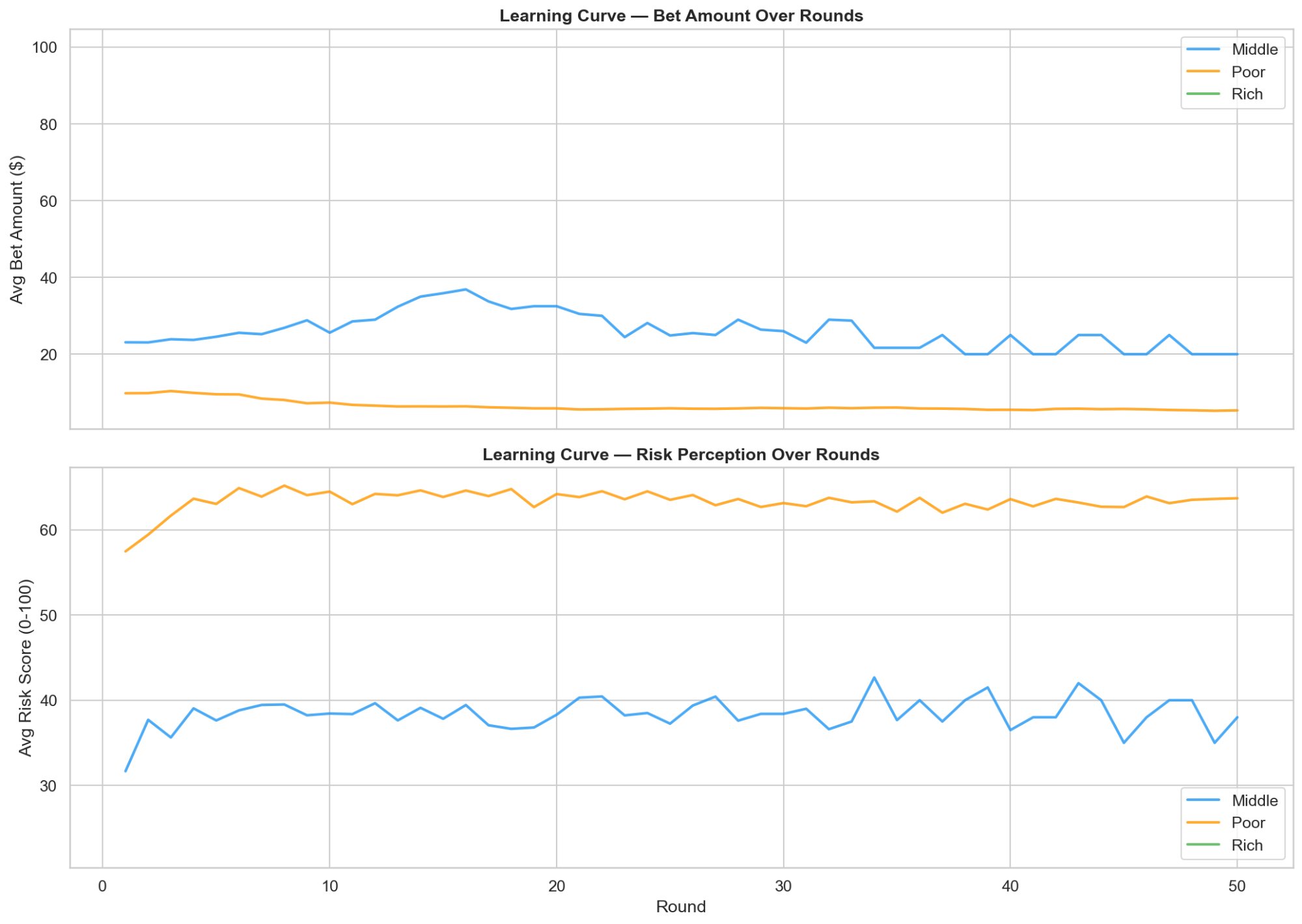}
  \caption{Learning Curves. Both bet amount and risk perception stabilize within approximately 5
  rounds for all personas. The Poor persona's risk score locks at approximately 65 and shows
  near-zero Spearman correlation with round number ($\rho=0.032$). The Rich persona's green line
  is absent after rounds 1--2 due to near-universal early stopping.}
  \label{fig:learning}
\end{figure}

The Poor persona's Spearman $\rho$ of 0.032 between round number and risk score is the number that most
directly answers the belief-updating question. With 5,589 observations, this correlation is technically
significant ($p=0.016$), but the effect size is negligible. Round 1 and round 50 produce essentially
identical risk score distributions for the Poor persona. The model enters each session with a risk
assessment anchored by its persona context and does not revise it meaningfully, regardless of what the
environment produces.

This is not a sampling artifact. The Middle persona shows a somewhat larger $\rho$ ($+0.320$),
reflecting a modest early rise in risk scores across the first five rounds before plateauing. This
suggests the model can respond to early trial outcomes but stops updating once it reaches an equilibrium
consistent with its persona prior. A genuine Bayesian learner would continue to revise its estimates as
evidence accumulates, particularly on the Biased Low machine where 65\% of rounds end in losses. The
model does not do this.

\subsection{Streak Effects: Hot Hand Sensitivity}

\begin{figure}[H]
  \centering
  \includegraphics[width=0.78\linewidth]{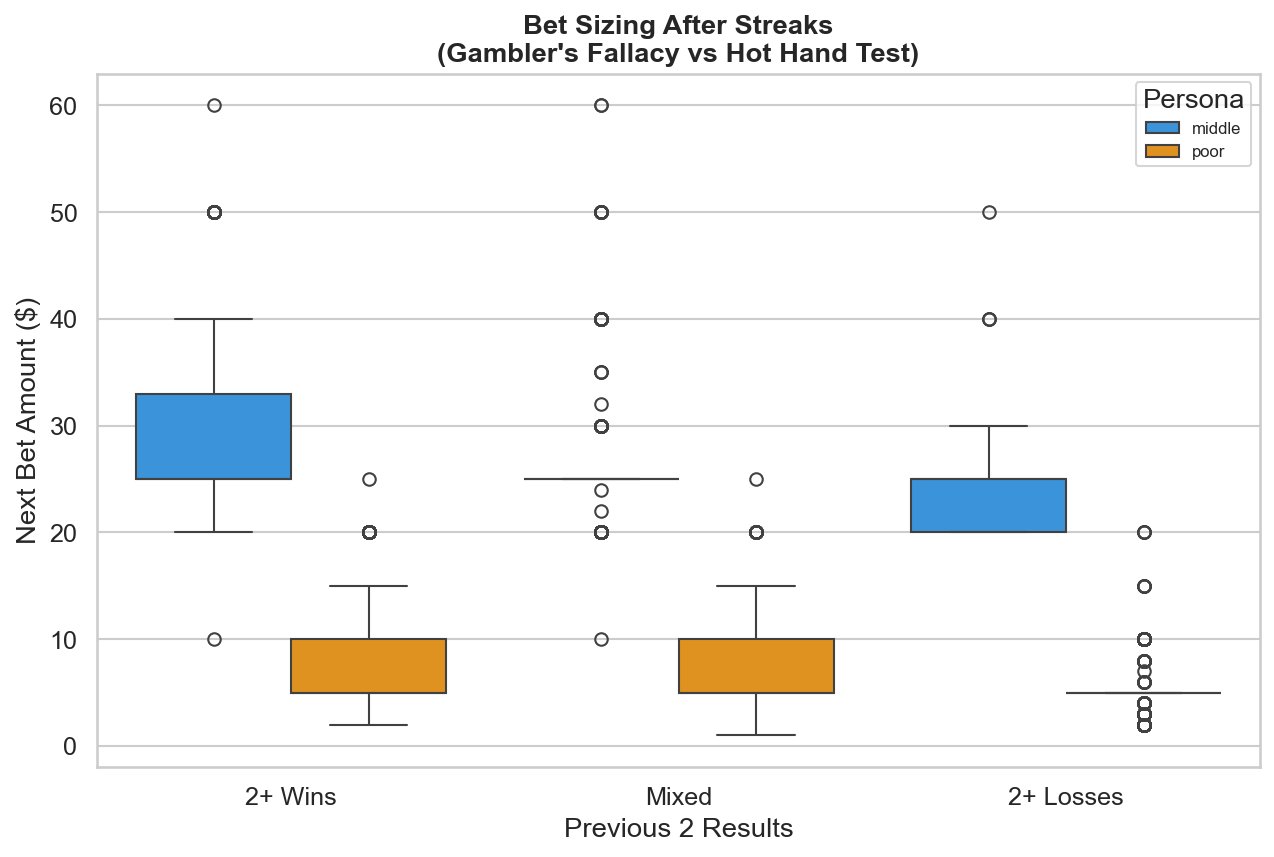}
  \caption{Bet Sizing After Streaks. The Middle persona bets approximately \$30 after 2$+$ wins
  versus \$25 after losses. The Poor persona shows stable low bets across conditions.}
  \label{fig:streak}
\end{figure}

The Middle persona demonstrates sensitivity to win streaks: mean bet after 2$+$ consecutive wins is
approximately \$30, compared to \$25 after losses, a 20\% increase that is visible in the median as
well. This is consistent with the Hot Hand Fallacy, interpreting a win streak as a positive signal
about future outcomes in an environment where the machines are stationary or randomly dynamic. The Poor
persona's bet stability across streak conditions is more consistent with resource conservation than with
streak-based inference: having fewer reserves, it cannot afford to vary its bet substantially regardless
of recent history.

\subsection{Win Rate and Return on Investment}

\begin{figure}[H]
  \centering
  \includegraphics[width=\linewidth]{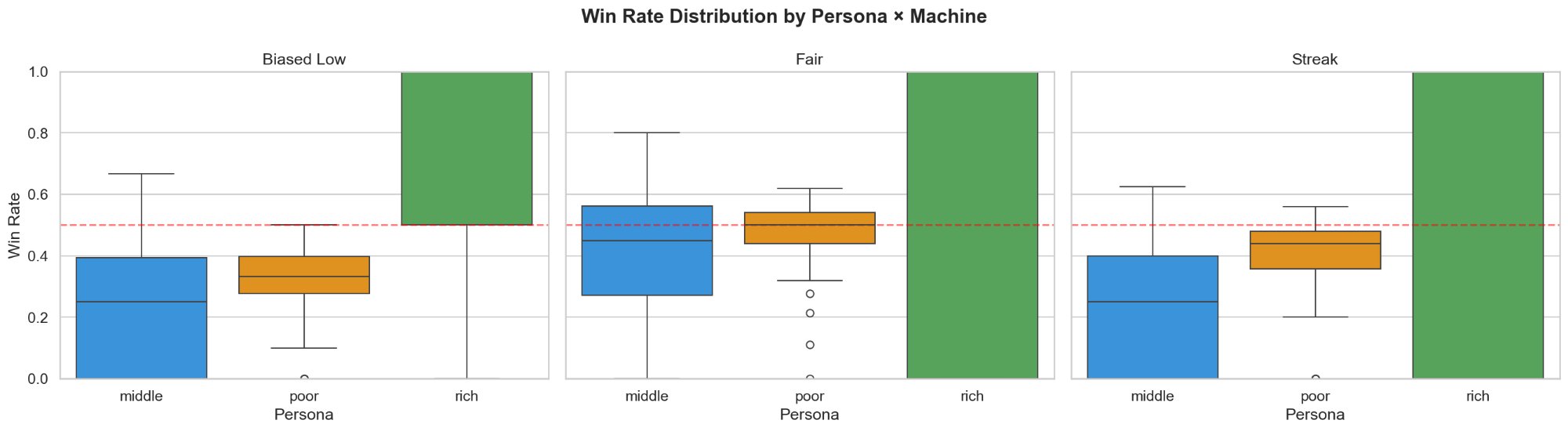}
  \caption{Win Rate Distribution by Persona and Machine Type. The Rich persona shows extreme
  variance due to near-zero session length (single-round sessions yield 0\% or 100\%). Middle
  and Poor converge toward true machine probabilities over longer sessions.}
  \label{fig:winrate}
\end{figure}

\begin{figure}[H]
  \centering
  \includegraphics[width=\linewidth]{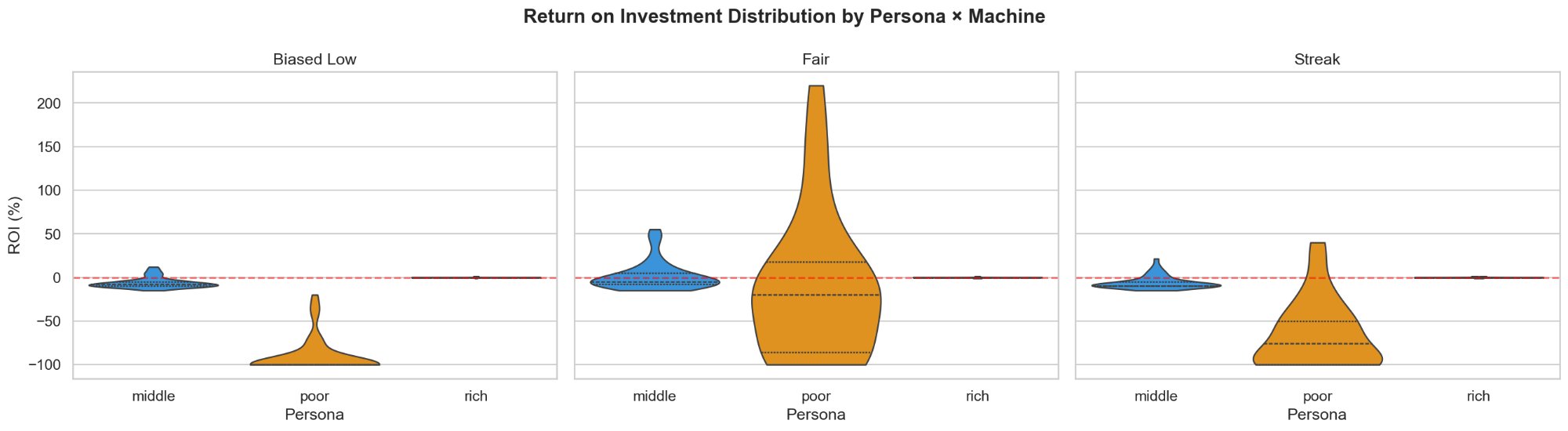}
  \caption{Return on Investment Distribution. Poor persona shows the widest ROI spread on the Fair
  machine, with a tail extending past 200\% in some sessions. Most Poor sessions end in moderate
  losses, but the long right tail reflects the high-variance nature of sustained risk-seeking play.}
  \label{fig:roi}
\end{figure}

The win rate data requires careful interpretation for the Rich persona. Because most Rich sessions are
single rounds, the win rate is either 0\% or 100\% by definition. The wide interquartile range is a
distributional artifact of sample size, not genuine performance variance. For the Middle and Poor
personas, win rates converge toward true machine probabilities over longer sessions, as expected. The
Streak machine produces modestly elevated win rates for the Poor persona compared to Biased Low, a quiet
confirmation that the dynamic probability structure is functional: patient play does eventually benefit
from the increasing win probability.

The ROI distributions confirm the high-variance character of the Poor persona's strategy. The Fair
machine produces the widest distribution, with some iterations yielding over 200\% return. For a persona
starting at \$50, a 200\% ROI means walking away with \$150, a tripling of wealth. Most sessions end at
moderate losses, but the existence of this tail is precisely what loss-domain risk-seeking predicts: the
Poor persona is effectively purchasing lottery-like variance, accepting frequent moderate losses for
occasional large gains.

\section{Discussion}

\subsection{Prospect Theory Reproduced Without Instruction}

The statistical evidence for persona-conditioned behavioral differentiation is, at this point, not in
question. Effect sizes of $r=1.000$ for session length and $d=4.15$ for risk score do not emerge from
noise. The question is what they mean.

One interpretation is that the model has internalized a representation of the behavioral economic value
function from its training data. Human text is saturated with descriptions of financial decision-making,
and Prospect Theory's predictions about how poor people and rich people respond to risk are well
documented in that text. The model may be drawing on these representations when instantiated in a
matching persona, producing behaviors that are consistent with the theory without explicitly reasoning
about it.

A more deflationary interpretation is that the model is performing stereotype completion: the prompts
invoke cultural representations of poor and wealthy people, and the model reproduces those
representations. These hypotheses are not easily distinguished with the current data. But they may not
be as different as they appear. If stereotype completion and formal economic theory predict the same
behaviors in the same direction and with the same relative magnitudes, then whether we call it theory or
stereotype may be a question of framing rather than mechanism.

\subsection{Emotions as Narration, Not Causation}

The chi-square result ($\chi^2=3205.4$, Cram\'{e}r's $V=0.392$) confirms a large association between
emotion and strategy labels. But the cell-level patterns reveal that this association is structured by
persona, not by causal logic. The \textsc{cautious} $+$ \textsc{risk\_seeking} pairing appears 302
times in a single condition. If caution caused risk aversion, this cell should be near-empty. It is not.

The most parsimonious explanation is that the model determines its behavioral decision based on persona
context and environmental history, and then generates emotional labels that are contextually coherent
with that decision. A risk-seeker can be cautious; the caution describes the approach, not the strategic
choice. This is recognizable human narration. The model is good at producing recognizable human
narration. The problem arises when we try to use those narratives as causal explanations of the model's
behavior.

This has a direct implication for interpretability research. Using an LLM's stated emotional or
motivational state as evidence about why it made a decision is unreliable if those states are post-hoc
annotations. The annotation may be accurate or inaccurate, and without independent probes of the
decision process, we cannot easily tell the difference.

\subsection{Belief Rigidity and Its Practical Consequences}

The near-zero Spearman correlation between round number and risk score for the Poor persona
($\rho=0.032$) is the finding we believe most warrants attention from practitioners. An LLM agent
deployed in a novel financial, medical, logistical, or adversarial environment will carry behavioral
priors encoded from its training data and initial prompt context. Our data suggests those priors will
dominate its behavior throughout an extended session, with only weak updating from sequential feedback.

This does not mean LLMs cannot update beliefs at all. The Reflexion framework~\cite{shinn2023} shows
that explicit reflection prompts can enable meaningful revision. What our data suggests is that implicit
belief updating from trial outcomes, without any architectural support for reflection, is weak.
Designing LLM agents for environments that require genuine adaptation likely requires explicit
mechanisms: structured reflection prompts, belief state tracking, or reinforcement from verified
outcomes, rather than relying on the base model to learn in-context.

\subsection{Limitations}

We acknowledge the following limitations:

\begin{itemize}
  \item \textbf{Single model:} All data comes from GPT-4.1. Generalizability to other models is unknown and is the most pressing gap.
  \item \textbf{Temperature:} The default temperature of 1.0 was used. The robustness of persona-driven behavioral patterns across temperature settings has not been tested.
  \item \textbf{Self-report reliability:} Psychological state scores are self-reported by the model. There is no guarantee they reflect genuine internal computational states rather than contextually plausible outputs.
  \item \textbf{No formal post-hoc correction for multiple comparisons across all tests:} Bonferroni correction was applied within the primary session-length comparisons. Exploratory analyses in later sections should be interpreted with appropriate caution.
  \item \textbf{Wealth confound:} Starting balances differ by a factor of 200. Future work should test intermediate wealth levels and proportionally scaled bet constraints to separate wealth-level effects from scale effects.
  \item \textbf{No cross-session memory:} The model had no memory between iterations. Long-term learning dynamics were not studied.
  \item \textbf{Explorer persona excluded:} The content filter exclusion removes what may have been the most scientifically interesting behavioral condition.
\end{itemize}

\section{Conclusion}

This paper set out to ask a narrow question about a language model's behavior in a gambling environment
and found answers that extend well beyond the specific setting. GPT-4.1, when asked to inhabit different
economic identities, behaves in ways that are recognizably human and specifically recognizable in the
terms that Kahneman and Tversky developed. The statistical evidence is strong: session length effect
sizes of $r=1.000$, risk score Cohen's $d$ of 4.15, and a consistent ordering across all conditions
and machine types.

At the same time, we found that the model's self-narration is not reliably causal. Emotional labels
appear decorative as often as diagnostic. Risk scores do not predict bet sizes in the direction
consistency demands. Internal beliefs update weakly from experience. The model is, in some ways, a very
persuasive narrator of its own behavior without being a particularly accurate one.

For researchers studying LLM cognition, this work provides a high-volume behavioral dataset, a framework
for connecting LLM outputs to established economic theory, and a set of statistical methods applicable
to future behavioral LLM experiments. For practitioners building LLM agents, the belief rigidity
finding is a practical caution: design explicit reflection mechanisms rather than assuming the model will
learn from sequential feedback. The behavioral prior set by an initial prompt will likely persist longer
than you expect.

\section{Future Work and Scalability}

\subsection{Reproducibility and Cross-Model Extension}

One of the quieter virtues of this experimental design is how little it asks of the next researcher who
wants to use it. The slot machine environment, the persona prompts, and the JSON response schema are all
self-contained. Any model that accepts a system prompt and returns structured output can be dropped into
the same protocol. A full replication of the study as described here requires roughly 6,950 API calls
per model. For open-source models run locally through frameworks like Ollama, the marginal cost is
electricity.

The most valuable immediate extension would be a comparative study across model families. The behavioral
measures reported in this paper --- session length separation, risk score differentiation, belief
rigidity across rounds, and emotion-strategy coherence --- form a concrete set of dependent variables
against which any model can be measured. Natural candidates include Claude 3.5 and 3.7 (Anthropic),
Gemini 1.5 Pro and 2.0 (Google DeepMind), Llama 3.1 at both 8B and 70B parameter scales (Meta,
open-source), Mistral Large, and Qwen 2.5. A reduced protocol of 20 iterations per condition rather
than 50 would yield over 1,200 decisions per model, sufficient for all primary comparisons reported here.

The present study used GPT-4.1 via Azure OpenAI, a mature and well-aligned proprietary model. Smaller
or less instruction-tuned models may show weaker persona conditioning, more erratic belief trajectories,
or qualitatively different emotion-strategy patterns. These differences would not be failures of
replication. They would be findings.

\subsection{A Socioeconomic Behavioral Index for LLMs}

A cross-model replication creates the conditions for something more structured: a standardized
Socioeconomic Behavioral Index, or SBI, that characterizes how faithfully a model reproduces human-like
economic behavior under persona conditioning. Based on the metrics developed in this study, we propose a
five-component index.

\textbf{Prospect Theory Alignment} captures how strongly a model differentiates session length across
the Rich and Poor personas, measured as the average Mann-Whitney effect size $r$ across the two primary
pairwise comparisons. GPT-4.1 scores 0.951 on this component, close to the theoretical maximum.

\textbf{Belief Rigidity} captures the absence of belief updating across rounds, measured as 1 minus the
absolute Spearman correlation between round number and risk score for the Poor persona. GPT-4.1 scores
0.968 here, which is a high score but not a flattering one. It means the model rarely revises its risk
assessment regardless of what the environment produces.

\textbf{Emotion-Decision Decoupling} captures how often emotional self-reports contradict the behavioral
decision they accompany, measured as the proportion of rounds where \textsc{cautious} co-occurs with
\textsc{risk\_seeking}. GPT-4.1 scores 0.61, meaning more than half of all cautious-labeled rounds are
paired with risk-seeking behavior.

\textbf{Environmental Sensitivity} captures how accurately the model distinguishes machines with
different true win probabilities, measured as the normalized fairness score gap between the Fair and
Biased Low machines. GPT-4.1 scores only 0.057 here, reflecting the limited discrimination documented
in Section~4.7.

\textbf{Persona Stability} captures how consistently a model maintains its behavioral identity across
independent runs of the same condition, measured as the inverse coefficient of variation of risk scores
within each persona.

These five components can be averaged into a single SBI score or displayed as a radar profile, enabling
direct visual comparison across models. The index is not a measure of general model quality. A high
belief rigidity score, to be clear, reflects a limitation rather than a strength. The SBI is a targeted
diagnostic for researchers and developers who need to understand how a specific model will behave when
given an economic identity and placed in a world that pushes back.

\subsection{Directions for Extension}

Several extensions would deepen the research program this paper opens. Adding intermediate wealth levels
between \$50 and \$10,000 would allow a finer mapping of the value function and test whether behavioral
transitions are monotonic or show unexpected non-linearities. Varying temperature systematically would
reveal whether persona-driven differentiation is robust to sampling stochasticity. Introducing a
cross-session memory mechanism, such as a brief summary of prior outcomes prepended to each new session,
would allow testing of longer-horizon belief updating, a question the current design cannot answer.

The Explorer persona remains an unfinished thread. The goal of that condition was to probe machine
behavior through deliberate, systematic experimentation rather than profit-seeking. It is, in many ways,
the most scientifically honest behavioral mode the experiment could have tested. Getting it past the
content filter is a puzzle worth solving.

The broader ambition is straightforward. Every model that gets tested adds a data point to a growing
picture of how language models, as a class, internalize and reproduce the economic reasoning of the
humans whose writing trained them. That picture is worth completing.


\end{document}